\documentclass[
]{ceurart}

\usepackage{tcolorbox}
\usepackage{graphicx}
\usepackage{multirow}
\usepackage[normalem]{ulem}
\usepackage{subcaption}
\useunder{\uline}{\ul}{}

\begin{document}

\copyrightyear{2021}
\copyrightclause{Copyright for this paper by its authors.
  Use permitted under Creative Commons License Attribution 4.0
  International (CC BY 4.0).}

\newtcolorbox{mybox}[1]{%
    colbacktitle=gray!30,
    coltitle=black,
    fonttitle=\bfseries,
    colback=gray!30,
    colframe=gray!10,
    sharp corners,
    title=#1}

\conference{HASOC (2021) Hate Speech and Offensive Content Identification in English and Indo-Aryan Languages}

\title{Classification on Sentence Embeddings for Legal Assistance}

\author[1]{Arka Mitra}[%
orcid=0000-0003-1071-7294,
email=thearkamitra@iitkgp.ac.in,
url=https://n.ethz.ch/~amitra/,
]
\address[1]{Indian Institute of Technology, Kharagpur, India}
\begin{abstract}
  Legal proceedings take plenty of time and also cost a lot. The lawyers have to do a lot of work in order to identify the different sections of prior cases and statutes. The paper tries to solve the first tasks in AILA2021 (Artificial Intelligence for Legal Assistance) that will be held in FIRE2021 (Forum for Information Retrieval Evaluation). The task is to semantically segment the document into different assigned one of the 7 predefined labels or "rhetorical roles." The paper uses BERT to obtain the sentence embeddings from a sentence, and then a linear classifier is used to output the final prediction. The experiments show that when more weightage is assigned to the class with the highest frequency, the results are better than those when more weightage is given to the class with a lower frequency. In task 1, the team legalNLP obtained a F1 score of 0.22.
\end{abstract}


\maketitle

\section{Introduction}
 Legal systems in many countries like USA, UK, Canada, India has two main sources- Precedents and Statutes; Precedents are previous similar cases while statutes are written laws that have to be followed in the country. The number of legal cases have been increasing and thus it is quite difficult for a lawyer to go through many of the precedents. Additionally, the legal reports in different countries are structured in different ways. Due to the lack of standardization, it is necessary to have a method that can help the lawyer to identify the different sentences in the report and process the report faster, while obtaining the relevant information quickly. The task 1 of AILA 2021 aims for the semantic segmentation of the document to assist the lawyer to process the information faster.\\
 AILA 2021, held in collocation with FIRE 2021, has several tasks for legal informatics. Legal documents follow certain sections like ``Facts of the Case", ``Issues being discussed" etc which are called ``rhetorical roles". For task 1, the sentences had to be classified into one of the seven different classes. More details on the classes and the dataset have been given in section \ref{sec:Dataset}.\\
 The remaining of the paper is divided into the following sections: \ref{sec:RelatedWork} which goes through the related work done for Rhetorical labelling in legal reports; \ref{sec:Dataset} that provides a detail of the dataset that has been used; \ref{sec:Method} describes the methodology that has been used for the paper; \ref{sec:results} showcases the results that have been obtained; \ref{sec:discussion} discusses the results and provides insights on the different models that have been used. \ref{sec:futureworks} talks of the future work that would be done and \ref{sec:conclusion} concludes the paper.
 
 \section{Related Work}
 \label{sec:RelatedWork}
    Text segmentation has been an important task in natural language processing. There have been probabilistic approaches that used Hidden Markov Models \cite{BorkarVinayak2001Asot} and Maximum Entropy Markov Models \cite{McCallum2000MaximumEM}. Saravanan and Ravindran \cite{saravanan-rhetoric} used Conditional Random Fields (CRF) for the identification of rhetoric labels for the segmentation and summarization of legal documents. Avelka et al. \cite{avelka2018SegmentingUC} used CRFs on annotated data from the US cyber crime and trade secrets decisions. Bhattacharya et al. \cite{Bhattacharya2019IdentificationOR} used CRF on top of a Bi-LSTM network to classify the sentences into different categories.\\
 \section{Dataset}
 \label{sec:Dataset}
    The AILA track started in 2019 \cite{Bhattacharya2019FIRE2A} and it had focused on Precedent and Statute retrieval. The second version of the same track focused \cite{fire2020-aila} on precedent and statute retrieval as well as rhetoric labelling \cite{bhattacharya2019identification}. The third iteration of the track \cite{fire2021-aila} also includes rhetoric labelling  but at the same time, it contains a task for automatic summarization \cite{parikh2021fire}.\\
    There are 60 documents in the task 1 dataset with 11285 labelled sentences. Each of the sentences has one of the seven possible labels:
    \begin{itemize}
        \item Facts : Sentences that discuss the facts about the case
        \item Ruling by Lower Court : The dataset contains Indian Supreme Court cases, which usually have a ruling at a lower court like High Court or Tribunal; the label indicates that the sentences are the decisions given in the lower court
        \item Argument : Arguments provided by the different parties
        \item Statute : The statute corresponding to the present case
        \item Precedent : The precedent corresponding to the present case
        \item Ratio of the decision : The reasoning given by the Supreme Court for the decision
        \item Ruling by Present Court: The final decision given by the Supreme Court
    \end{itemize}
\section{Method}
 \label{sec:Method}
 The first subsection discusses the approach for the task and the next subsection provides the experimental details.
 \subsection{Approach}
 The dataset contained seven different classes but the distribution among those classes are quite skewed. Table \ref{tab:data_desc} shows that the number of samples with the label ``Ratio of decision" is about 12 times the number of samples with the label ``Ruling by Present Count". It is important to keep the number of samples almost equal- that would allow the model to learn meaningful information from each of the classes. \\
 \begin{figure}
     \centering
     \includegraphics[width=0.3\columnwidth]{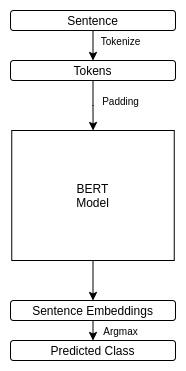}
     \caption{Overall Methodology}
     \label{fig:overall}
 \end{figure}
 There are three main methods to do that. In the first method, the sampling from the dataset can be done in such a way that more samples from the lower frequency class are taken and lower samples are taken from the class with higher frequency. The downside to this method is that we are leaving several samples from the dataset which would decrease the performance of the model. In another method, one can do the sampling in such a way that the same example of the class with lower frequency is included into the batch multiple times such that the total distribution in the new dataset has the same number for each of the classes. However, since the same example is chosen multiple times, it is increasing the chances of overfitting and also increases the computation overhead. The last method keeps the computational cost about the same as the first method but at the same time does not change the dataset size. In this method, the loss is changed so that false predictions for the class with lower number of samples is penalized more. The new loss, as shown in Eqn. \ref{eqn:modce}, is a modified version of the cross-entropy loss. The weight multiplied with the cross-entropy can be considered as the number of times the sample has been considered. A class with a higher number of samples in the dataset needs to have a lower weight associated with it.\\
 \begin{equation}
\label{eqn:modce}
    loss(logits, class) = weight[class]*(-log(\frac{exp(logits[class])}{\sum_j exp(logits[j])}))
\end{equation}
 The author prepossessed the document and combined the sentences in all the documents and the associated labels to create the dataset that has been used. BERT\cite{Devlin2019BERTPO} was used for creating the sentence embeddings for the sentences. BERT has been trained on a lot of data and thus would be able to create a condensed representation of the sentence. The output of the ``CLS" token from the BERT output was considered to be the sentence embedding for the sentence and then that was sent through a linear layer. The logits obtained from the linear layer were considered and the maximum of them was selected to be the predicted class of the section. The overall methodology has been described in Figure \ref{fig:overall}.
 
 \begin{table}
    \caption{The distribution of the different classes}
    \label{tab:data_desc}
    \begin{tabular}{|c|c|}
    \hline Label & Count of Label \\
    \hline Ratio of the decision & 4211 \\
    \hline Facts & 2622 \\
    \hline Precedent & 1787\\
    \hline Argument & 939 \\
    \hline Statute & 902 \\
    \hline Ruling by Lower Court & 483\\
    \hline Ruling by Present Court & 341\\
    \hline
\end{tabular}
\end{table}
\subsection{Experimental Details}
The cased and uncased BERT model have been implemeneted with the help of the Huggingface library \cite{wolf2020huggingfaces}. Pytorch has been used for the framework. The batch size of 8 has been considered and the model has been trained for 4 epochs. 80\% of the data has been considered as the training set and the rest is considered as the validation set. The model weights which gave the best results on the validation set had been saved and used for inference on the test set. AdamW optimizer \cite{Loshchilov2019DecoupledWD} with an initial learning rate of 2e-5 is used for training. The max length for the padding was kept at about 0.98 percentile which is around 120 tokens. The codes are publicly available on github\footnote{https://github.com/thearkamitra/LegalNLP}. The random seed has been set to 42.
 \section{Results}
 \label{sec:results}
 There are three runs which have been submitted finally. The macro-F1 score, precision and recall of the different runs are given in Table. \ref{tab:results}.\\
 \begin{table}[h!]
    \caption{The results of the different runs}
    \label{tab:results}
    \begin{tabular}{|c|c|c|c|}
    \hline Run number & Precision & Recall & F1\\
    \hline legalNLP\_1 & 0.197 & 0.217 & 0.196 \\
    \hline legalNLP\_2 & 0.198 & 0.215 & 0.192 \\
    \hline legalNLP\_3 & 0.225 & 0.227 & 0.22\\
    \hline
\end{tabular}
\end{table}
The description of the three runs are as follows:
\begin{itemize}
    \item The first run uses base cased BERT with weights to modify the cross entropy loss
    \item The second run uses base uncased BERT with the same weights as the previous run.
    \item The third and final run uses base cased BERT but here the weights are inverted such that the class with higher number of samples is given more weights.
\end{itemize}
\section{Discussion}
 \label{sec:discussion}
 The results show that the cased BERT model performed better than the uncased model. This can be explained by the fact that there may be some phrases in legal reports that have different meanings when used in uppercase vs lowercase and the BERT cased model is able to capture the contextual information. Due to the better performance of cased BERT, the author performed the same experiment with the same random seeds but with different weight for the cross-entropy loss. Comparison between the first and third run shows that the model performed better when more weightage was given to classes that existed more abundantly. This contradicts the belief that a model trained with skewed distribution would perform worse than one without. A possible explanation might be that the test data has more sentences with labels corresponding to the higher class. As a consequence, the metric reports a higher score for the third metric.\\
 \section{Future Work}
 \label{sec:futureworks}
 In the present work, the sentences from the documents had been extracted and aggregated to form the dataset. However, there is a relation between the labels and where the sentence is located in the document (for example, Ruling by Present Court is always present in the final ending of the documents). Also, the author has not considered the co-occurance of the different label. For that, Hidden markov model or some probabilistic state machine could be used to further improve the accuracy of the model. 
 \section{Conclusion}
 \label{sec:conclusion}
 The paper describes the modified cross-entropy loss and the use of BERT models for rhetoric role labelling in legal documents. The three runs that had been submitted obtained a score of 0.196, 0.192 and 0.22 respectively.
\begin{acknowledgments}
The author thanks the organizers of Artificial Intelligence for Legal Assistance for creating this task. The author would also like to acknowledge Google Colab for providing the computational resources needed. The BERT model is built on the library made by Huggingface \cite{wolf2020huggingfaces}.
\end{acknowledgments}
\bibliography{w-ce}
\end{document}